\documentclass[letterpaper, 10 pt, conference]{ieeeconf}  

\IEEEoverridecommandlockouts                              

\usepackage{amssymb}
\usepackage{amsmath}
\usepackage{dsfont}
\usepackage{subcaption}
\usepackage{graphics} 

\PassOptionsToPackage{hyphens}{url}\usepackage[breaklinks=true,hidelinks=true,colorlinks=true,citecolor=blue,linkcolor=blue]{hyperref}

\usepackage{algorithm}
\usepackage{algpseudocode}
\usepackage{bm}
\usepackage{enumerate}
\usepackage{graphicx}
\usepackage{lipsum}
\usepackage{multirow}
\usepackage{tabularx}
\usepackage{framed}
\usepackage[dvipsnames]{xcolor}

\usepackage{pgfplots}
\usepackage{tikz}
\usepackage{booktabs}
\usetikzlibrary{positioning}
\usetikzlibrary{calc, arrows, fit,spy}
\usepgfplotslibrary{groupplots}

\graphicspath{{images/}}

\makeatletter

\DeclareMathOperator*{\argmaxB}{argmax}

\newcolumntype{Y}{>{\centering\arraybackslash}X}

\newcommand{\ignore}[1]{}

\title{\LARGE \textbf{CropNeRF: A Neural Radiance Field-Based\\ Framework for
Crop Counting}}

\author{Md Ahmed Al Muzaddid and William J. Beksi
\thanks{The authors are with the Department of Computer Science and
        Engineering, The University of Texas at Arlington, Arlington, TX, USA.
        Emails:
        mdahmedal.muzaddid@mavs.uta.edu,
        william.beksi@uta.edu.
        }
}

\begin{document}

\maketitle
\pagestyle{plain}

\begin{abstract}
Rigorous crop counting is crucial for effective agricultural management and
informed intervention strategies. However, in outdoor field environments,
partial occlusions combined with inherent ambiguity in distinguishing clustered
crops from individual viewpoints poses an immense challenge for image-based
segmentation methods. To address these problems, we introduce a novel crop
counting framework designed for exact enumeration via 3D instance segmentation.
Our approach utilizes 2D images captured from multiple viewpoints and associates
independent instance masks for neural radiance field (NeRF) view synthesis. We
introduce crop visibility and mask consistency scores, which are incorporated
alongside 3D information from a NeRF model. This results in an effective
segmentation of crop instances in 3D and highly-accurate crop counts.
Furthermore, our method eliminates the dependence on crop-specific parameter
tuning. We validate our framework on three agricultural datasets consisting of
cotton bolls, apples, and pears, and demonstrate consistent counting performance
despite major variations in crop color, shape, and size. A comparative analysis
against the state of the art highlights superior performance on crop counting
tasks. Lastly, we contribute a cotton plant dataset to advance further research
on this topic. 
\end{abstract}


\section{Introduction}
\label{sec:introduction}
With a declining migrant workforce and a growing global population, the
agricultural sector faces significant challenges that demand innovative
solutions. Moreover, the adverse effects of climate change, such as
unpredictable weather patterns and water scarcity, necessitate the creation of
more adaptive and efficient farming practices. By leveraging advanced
technologies (e.g., robotic automation, computer vision, sensor networks, etc.)
to improve resource efficiency and optimize productivity, precision agriculture
is a practical and effective response to these complications.

Accurate crop counting is essential for resource optimization, yield estimation,
and post-harvest management in precision agriculture. Nonetheless, counting
crops in field conditions remains problematic. For example, occlusions caused by
foliage, branches, and neighboring plants complicates crop detection and
re-identification thus increasing the likelihood of double counting. In
addition, crop overlap and clustering hinder the precise differentiation of
individual instances, while variations in color, shape, and size across growth
stages further impede the counting process.

\begin{figure}
\centering
\hspace{-13px}
\resizebox{\linewidth}{6cm}{\input{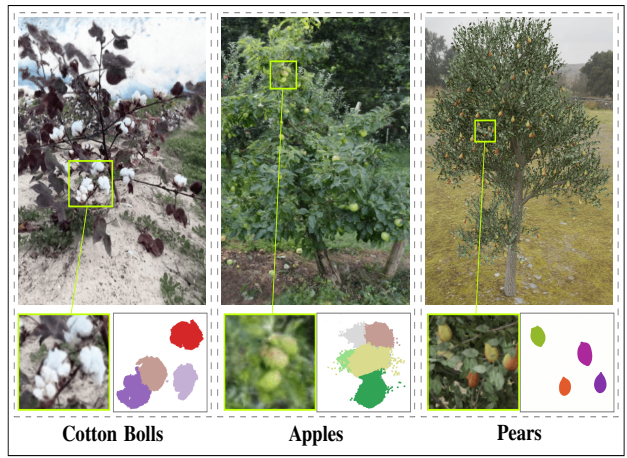}}
\caption{An overview of crop counting via 3D instance segmentation. The
zoomed-in views depict clusters of cotton bolls, apples, and pears from a
cotton plant, apple tree, and pear tree, respectively. The instance
segmentation results for these clusters, obtained using CropNeRF, are shown in
color on the right-hand side of the zoomed-in views.}
\label{fig:overview}
\vspace{-2mm}
\end{figure}

Various approaches have been proposed to address these challenges
\cite{he2022fruit}, yet very few leverage the 3D structure of the problem. In
this paper, we introduce a novel approach to instance segmentation that
effectively distinguishes individual instances within a reconstructed 3D
environment, thereby providing an accurate enumeration of the target crop.  Our
method demonstrates resilience to inaccuracies in the underlying 2D instance
segmentation masks upon which it depends. Additionally, it exploits volumetric
information to handle issues such as occlusion and perspective-induced
ambiguity, and can be deployed on crops of different physical qualities minus
the need for recalibration.

This work presents a crop neural radiance field (CropNeRF) 3D instance
segmentation framework that accurately differentiates individual instances of
target crops, Fig.~\ref{fig:overview}. By integrating visibility and instance
mask consistency scores, CropNeRF effectively addresses occlusions and
ambiguities caused by changes in perspective while adapting to crops of varying
dimensions without requiring sensitive parameter tuning. Unlike existing
instance segmentation approaches (e.g., \cite{liu2023instance,ye2024gaussian}),
which depend on establishing correspondences between instance masks across
multiple images, CropNeRF is designed to operate independently on each mask.
This eliminates the need for explicit inter-image mapping, thereby simplifying
the 3D segmentation process while improving accuracy.

\begin{figure*}
\centering
\resizebox{.99\linewidth}{!}{\input{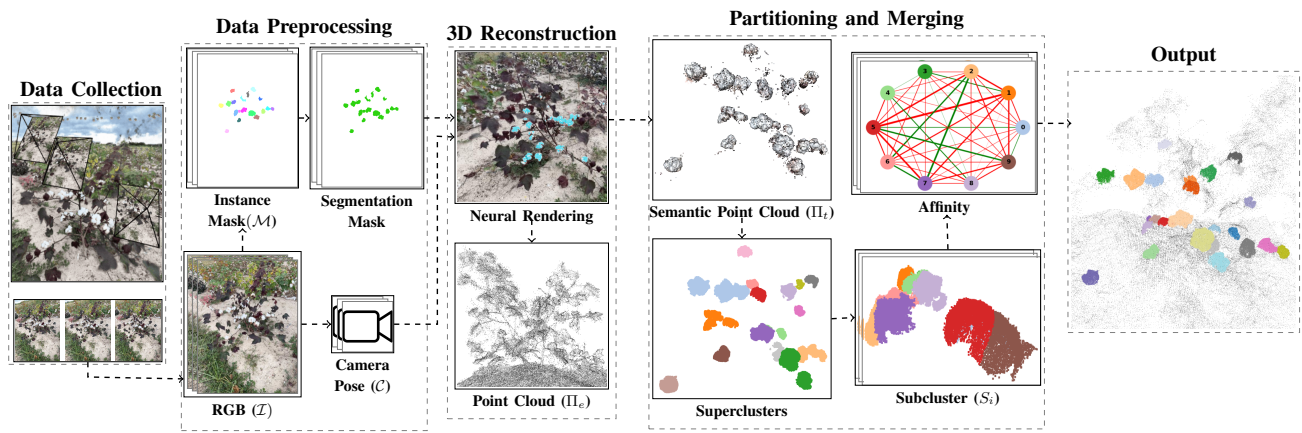}}
\caption{The crop counting pipeline. Multiple images of the target crop are
captured from different viewpoints during the data collection step. In the data
preprocessing stage, camera poses ($\mathcal{C}$) are extracted from the
captured images ($\mathcal{I}$), and the crops are segmented into instance masks
($\mathcal{M}$). Semantic segmentation masks are then generated from the
instance masks. The extracted camera poses, captured images, and semantic masks
are used to train a semantic NeRF model during the 3D reconstruction stage. The
trained model generates a semantic point cloud representing the crops ($\Pi_t$)
and a point cloud representing the environment ($\Pi_e$). In the partitioning
and merging step, the semantic point cloud is first divided into superclusters
and then into subclusters ($S_i$). The affinity among subclusters is calculated
based on subcluster visibility and mask consistency scores. Lastly, the
subclusters are merged based on subcluster affinity to produce the final
output.}
\label{fig:cropnerf_pipeline}
\vspace{-3mm}
\end{figure*}

Given a set of images from multiple viewpoints along with the corresponding
camera poses and 2D instance masks of the target crop, CropNeRF accurately
counts instances of the crop via 3D instance segmentation. To do this, we learn
a high-fidelity 3D reconstruction while simultaneously modeling a semantic
field. From the semantic field, we extract a point cloud representing only the
targeted crop, which is then partitioned into subclusters. The visibility and
mask consistency of the subclusters is computed based on the camera poses,
corresponding point clouds, and associated instance masks. These scores are
then aggregated across multiple views, to merge the subclusters into distinct
instances of the target crop, allowing for exact crop counting.

In summary, we make the following contributions.
\begin{itemize}
  \item We develop a 3D instance segmentation method capable of differentiating
  crops of varying colors, shapes, and sizes.
  \item We introduce a mask reliability score that incorporates crop visibility
  and mask consistency, enabling robustness against occlusions and annotation
  discrepancies.
  \item We release a public infield cotton plant dataset designed for 3D
  rendering and cotton boll counting tasks.
\end{itemize}
The source code, dataset, and multimedia material associated with this project
can be found at
\href{https://robotic-vision-lab.github.io/cropnerf}{https://robotic-vision-lab.github.io/cropnerf}.

\section{Related Work}
\label{sec:related_work}
\subsection{Image-Based Techniques}
\label{subsec:image-based_techniques}
Image-based methods typically employ object detection to identify crops within
images. For example, Chen et al. \cite{chen2017counting} utilized multiple
convolutional neural networks (CNNs) to map input images to total fruit counts.
Similarly, H{\"a}ni et al. \cite{hani2018apple} formulated crop counting as a
multi-class classification problem where a CNN predicts a single count per
detected crop cluster. Tedesco et al. \cite{tedesco2020convolutional} applied
detection models (e.g., Faster R-CNN \cite{ren2016faster}, SSD
\cite{liu2016ssd}, and MobileNetV2 \cite{sandler2018mobilenetv2}) and selected
architectures based on computational constraints. Likewise, James et al.
\cite{james2024citdet,james2024few} evaluated object detection models along
with transfer learning to classify citrus fruit on trees versus on the ground
for yield estimation. While modern object detection models can achieve high
accuracy under outdoor lighting conditions, performance is often adversely
affected by occlusions due to limited perspective and the absence of depth
information.

\subsection{Video-Based Tracking Methods}
\label{subsec:video-based_tracking_methods}
To handle occlusions and facilitate farm-level crop counting, previous works
have implemented multi-frame techniques based on a tracking-by-detection
paradigm. For instance, Smitt et al. \cite{smitt2021pathobot} counted sweet
peppers in RGBD videos by applying Mask R-CNN \cite{he2017mask} for
segmentation along with wheel odometry and depth information to enhance
tracking accuracy. Zhang et al. \cite{zhang2022deep} modified YOLOv3
\cite{redmon2018yolov3} to improve the detection of small fruits and
implemented a region-based counting strategy for tracking. Al Muzaddid and
Beksi \cite{muzaddid2024ntrack} created a cotton boll counting framework that
leverages the spatial relationships among neighboring bolls to re-identify
occluded instances. In proceeding work, Al Muzaddid et al.
\cite{muzaddid2025croptrack} developed a crop tracking approach based on the
combination of appearance and motion information. Although these methods are
effective, their performance is dependent on the accuracy of the object
detectors and trackers.

\subsection{3D Reconstruction-Based Solutions}
\label{subsec:3d_reconstruction-based_solutions}
Structure-from-motion (SfM) has been employed to generate 3D point cloud
reconstructions of orchards for precise crop localization. A cotton boll
counting method by Sun et al. \cite{sun2020three} made use of 3D point clouds
reconstructed from multi-view field images using SfM. Matos et al.
\cite{matos2024tracking} proposed a frame-to-frame tracking technique for fruit
counting across image sequences by leveraging 3D information derived from SfM.
Nonetheless, these approaches often require extensive camera calibration and
may produce noisy reconstructions in environments with dense foliage.

NeRFs \cite{mildenhall2021nerf} are a popular methodology for 3D
reconstruction. They offer high-fidelity volumetric representations without
the need for extensive camera calibration. For example, FruitNeRF
\cite{meyer2024fruitnerf} is a NeRF-based fruit counting framework. It relies
on clustering-based segmentation, which assumes consistent crop sizes and
requires manual parameter tuning. This limits the generalizability of FruitNeRF
across different crop growth stages. In contrast, our approach eliminates the
need for delicate parameter tuning based on crop size and distribution.

3D Gaussian splatting (3DGS) has emerged as alternative to NeRFs. Jiang et al.
\cite{jiang2025cotton3dgaussians} used 3DGS to reconstruct high-fidelity 3D
models of cotton plants. However, the dataset was collected indoors, under
controlled lighting conditions, and includes only defoliated cotton plants.
Zhang et al. \cite{zhang2025wheat3dgs} applied 3DGS to reconstruct wheat plots
and extract morphological traits (e.g., head length, width, and volume) via
segmentation. Yet, the segmentation technique relies on an iterative process
based on the overlap between projected 3D Gaussians and 2D segmentation masks,
quantified using the intersection over union and filtered by empirical
thresholds. Differently, our method accounts for crop visibility when
evaluating mask overlap and eliminates the need for arbitrary thresholding. The
end result is a more robust, crop-agnostic segmentation framework suitable for
field conditions.

\section{Neural Radiance Field-Based Framework for Crop Counting}
\label{sec:neural_radiance_field-based_framework_for_crop_counting}
Let $\mathcal{I} =\{I_1, I_2,\ldots,I_n\}$ be a set of RGB images that capture
the target environment from multiple viewpoints, where $I_i\in
\mathbb{R}^{h\times w \times 3}$ and each image is of resolution $h \times w$.
In addition, assume that for each image an instance mask $\mathcal{M} =\{M_1,
M_2,\ldots,M_n\}$ of the target crop is available and $M_j\in
\mathbb{R}^{h\times w}$. We proceed by extracting camera poses $\mathcal{C}
=\{C_1, C_2,\ldots,C_n\} \in \mathbb{R}^{3 \times 4}$ from the set of images.
For consistency, we use $C_j$ and $M_j$ to denote the camera pose and instance
mask, respectively, associated with image $I_j$. Note that $C_j$ may be used
interchangeably to refer to the $j^{th}$ camera, its pose, or the associated
camera view where the context clarifies the intended meaning. An overview of the
crop counting pipeline is illustrated in Fig.~\ref{fig:cropnerf_pipeline}.

\subsection{Volumetric Rendering}
\label{subsec:volumetric_rendering}
Given the images $\mathcal{I}$ and their corresponding camera poses
$\mathcal{C}$, we train a NeRF model to reconstruct a high-fidelity 3D
representation of the environment. The model encodes the scene within a
multilayer perceptron via mapping a position $x = (x, y, z)$ and a viewing
direction $d = (\phi,\theta)$ to a volume density $\sigma$ and RGB radiance $c =
(r, g, b)$. The density field $\mathcal{F}_{\sigma} : x \rightarrow \sigma$ is
defined as a function of the position, while the appearance field $\mathcal{F}_c
: (x,d) \rightarrow c$ depends on both the position and the viewing direction.
Building upon the concept of semantic fields \cite{zhi2021place}, we augment the
NeRF framework to integrate crop-specific information. Concretely, we utilize a
set of masks $\mathcal{M}$ to embed crop information into the 3D volume through
a semantic field $\mathcal{F}_s : x \rightarrow s$ that predicts the spatial
logits.

\subsection{Point Cloud Generation}
\label{subsec:point_cloud_generation}
The trained NeRF model encodes the spatial information of the entire scene
(i.e., soil, trunk, stem, foliage, and crop) within the density field
$\mathcal{F}_{\sigma}$. Conversely, the semantic field $\mathcal{F}_s$
exclusively represents the spatial distribution of the target crop within the 3D
volume. From these two fields, we generate two separate point clouds, one
representing the environment (including the target crop) and another one
representing only the target crop, by sampling $\mathcal{F}_\sigma$ and
$\mathcal{F}_s$, respectively. To extract the point cloud representing the
environment $\Pi_e$, we uniformly sample points from $\mathcal{F}_\sigma$.
Nevertheless, extracting points that exclusively represent the target crop
$\Pi_t$ by sampling $\mathcal{F}_s$ leads to a sparse point cloud. To address
this issue, we follow the approach used in FruitNeRF where sampled semantic
points are filtered based on their corresponding density values resulting in a
denser representation.

\subsection{Point Cloud Preprocessing}
\label{subsec:point_cloud_processing}
\begin{figure}
\centering
\begin{minipage}{0.55\linewidth}
  \centering
  \includegraphics[width=\textwidth, height=7cm]{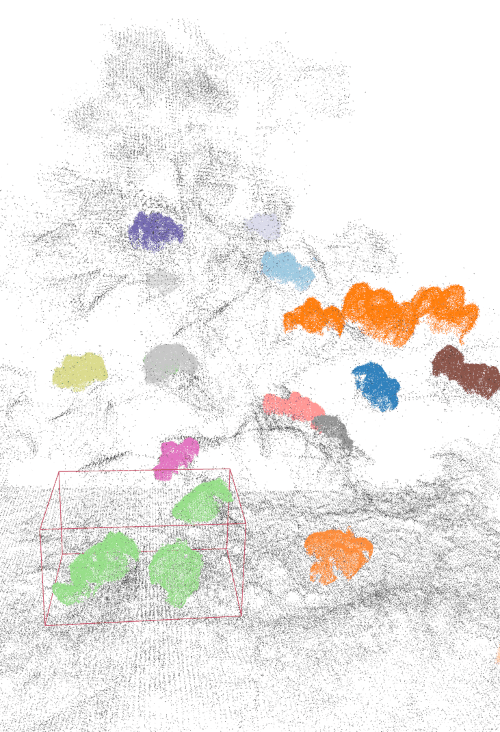}
  \subcaption{}
  \label{fig:multiple_superclusters}
\end{minipage}%
\hfill
\begin{minipage}{0.4\linewidth}
  \centering
  \includegraphics[width=\textwidth,trim={0cm 0cm .2cm 0cm},clip]{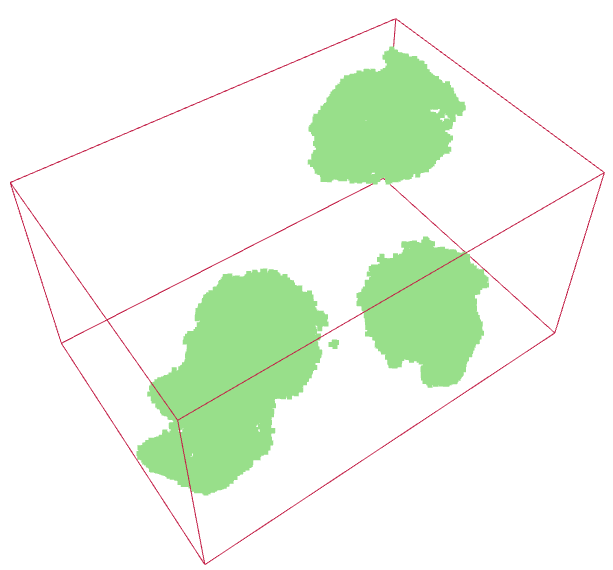}
  \subcaption{}
  \label{fig:single_supercluster}
  \vspace{5pt}
  \includegraphics[width=\textwidth]{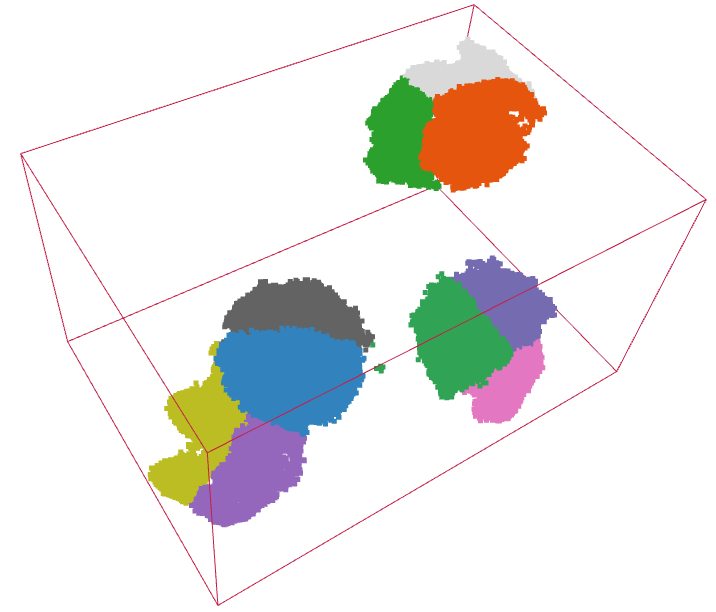}
  \subcaption{}
  \label{fig:multiple_subclusters}
\end{minipage}
\caption{The point cloud clustering process: (a) point cloud $\Pi_t$
representing the crops is segmented into superclusters, each identified by a
unique color; (b) a single supercluster visualized in 3D; (c) the supercluster
is further partitioned into multiple subclusters.}
\label{fig:clustering_process}
\end{figure}

We preprocess the point cloud $\Pi_t$ by partitioning it using a two-phase
approach. In the first phase, $\Pi_t$ is segmented into superclusters,
$\psi_1,\psi_2,\ldots,\psi_\kappa$, such that $\Pi_t
=\dot{\bigcup}_{k=1}^{\kappa} \psi_k$ where $\dot{\bigcup}$ denotes the disjoint
set union (Fig.~\ref{fig:multiple_superclusters}). The goal of this step is to
separate spatially distant and independent crop instances, thereby decomposing
the problem into smaller and more manageable subproblems that can be efficiently
addressed. We employ DBSCAN \cite{ester1996density} to identify the
superclusters based on point density without requiring prior assumptions about
the number or size of the clusters.

We refine the extracted superclusters by removing outliers based on the
clustering results, ensuring a more coherent segmentation
(Fig.~\ref{fig:single_supercluster}). It is important to observe that the output
of our method remains invariant to the number of superclusters. For example,
$\Pi_t$ can be considered as a single supercluster without affecting the final
segmentation results. However, to improve computational efficiency and minimize
unnecessary interactions between independent regions of the point cloud, we
incorporate the first-phase clustering as a preprocessing step. Each
supercluster is segmented independently and the results from all superclusters
are aggregated to generate the output.

In the second phase, we further divide a supercluster $\psi_i$ into $K$
subclusters. This is done using the k-means clustering algorithm, depicted in
Fig.~\ref{fig:multiple_subclusters}, such that $\psi_i =\dot{\bigcup}_{k \in K}
S_k\;$. The objective of this step is to partition a supercluster into smaller
subclusters, guaranteeing that each crop instance comprises at least one or more
subclusters. We achieve this by decreasing the size of each cluster through the
selection of a larger $K$ value. Note that for a moderately large $K$, the
segmentation results remain robust to variations in $K$ as discussed in the
evaluation (Sec.~\ref{subsec:k_vs_count}).

\subsection{Subcluster Visibility}
\label{subsec:subcluster_visibility}
Due to occlusions, some subclusters may only be partially observable from
certain viewpoints. To quantify their visibility, we assign a score $v_{ij} \in
[0,1]$ to each subcluster $S_i$ as viewed from camera $\mathcal{C}_j$. More
formally,
\begin{align}
  v_{ij} &=\frac{Area\;  of\;  S_i\; visible\; from\; \mathcal{C}_j }
                {Area\; of\; S_i\; visible\; from\; \mathcal{C}_j\; without\; occlusions} \nonumber\\
         &=\frac{Area\left(\mathcal{V}_j\left(S_i\right)\right)}{Area\left(\mathcal{P}_j\left(S_i\right)\right)}.
  \label{eq:visibility_score}
\end{align}
In \eqref{eq:visibility_score}, $\mathcal{P}_j(S_i)$ is a function that projects
point cloud $S_i$ onto the camera plane using the camera matrix, which is the
product of the intrinsic and extrinsic matrices derived from $\mathcal{C}_j$. We
refer to this as the occlusion-free projection of $S_i$. The occlusion-aware
projection function is represented by $\mathcal{V}_j(S_i)$ and accounts for
environmental elements such as the stems, foliage, and crops (i.e., any objects
between the camera $\mathcal{C}_j$ and $S_i$ that may occlude $S_i$). We obtain
$\mathcal{V}_j(S_i)$ by projecting both $S_i$ and $\Pi_e$ onto the camera plane
while incorporating depth information to account for occlusions.

\subsection{Mask Consistency}
\label{subsec:mask_consistency}
The instance masks $\mathcal{M}$ are based on 2D images and therefore inherently
susceptible to inconsistencies, regardless of whether they are annotated by a
machine learning model or a human. These deviations arise from the fixed camera
viewpoint and the absence of depth information, hindering precise
differentiation of closely-spaced instances. To account for variations in the
masks, we define a consistency score $c_{ij} \in [0,1]$. This score
quantitatively evaluates the consistency of the mask $M_j$ that overlaps with
$\mathcal{V}_j\left(S_i\right)$, the occlusion-aware projection of $S_i$. The
consistency score is computed as
\begin{equation}
  c_{ij} = \frac{\max\limits_l Area\, ( \mathcal{V}_j\left(S_i\right) \cap M_{jl})}{Area\left(\mathcal{V}_j\left(S_i\right)\right)},
  \label{eq:consistency_score}
\end{equation}
where $M_{jl}$ denotes the region of $M_j$ labeled as instance $l$. The $\max$
operation ensures that if multiple distinct labels in $M_j$ overlap with
$\mathcal{V}_j\left(S_i\right)$, then only the region with label $l$ that has
maximum overlap with $\mathcal{V}_j\left(S_i\right)$ is considered. We also
assign the label,
\begin{equation}
  \lambda_{ij} = \argmaxB_{\l} Area\, (\mathcal{V}_j\left(S_i\right) \cap M_{jl}),
\end{equation}
to subcluster $S_i$ in $M_j$.

\subsection{Mask Reliability}
\label{subsec:mask_reliability}
\begin{figure}
\centering
\resizebox{.95\linewidth}{!}{\input{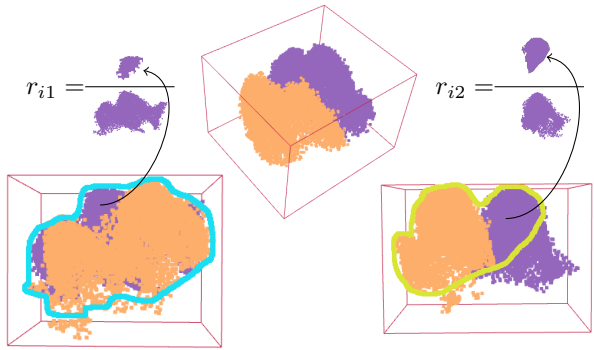}}
\caption{A visual representation of computing the mask reliability score. The
score of the purple subcluster is calculated based on two different camera
views. Only the visible projection area of the subcluster that overlaps with the
masks (represented by the cyan and yellow boundaries) is considered in the
numerator.}
\label{fig:mask_reliability}
\end{figure}

To quantify the confidence level of an instance mask associated with subcluster
$S_i$ in view $M_j$, we derive a reliability score $r_{ij} \in [0,1]$. As
illustrated in Fig.~\ref{fig:mask_reliability}, $r_{ij}$ is computed by
combining the subcluster visibility score $v_{ij}$ and the mask consistency
score $c_{ij}$. Concretely,
\begin{align}
  r_{ij} &=  v_{ij}c_{ij} \nonumber\\
         &= \frac{Area\left(\mathcal{V}_j\left(S_i\right)\right)}{Area\left(\mathcal{P}_j\left(S_i\right)\right)}
            \cdot
            \frac{\max\limits_l Area\, (\mathcal{V}_j\left(S_i\right) \cap M_{jl})}{Area\left(\mathcal{V}_j\left(S_i\right)\right)} \nonumber\\
         &=\frac{\max\limits_l Area\, ( \mathcal{V}_j\left(S_i\right) \cap M_{jl})}{Area\left(\mathcal{P}_j\left(S_i\right)\right)}.
\end{align}

\subsection{Merging Process}
\label{subsec:merging_process}
\begin{figure}
\centering
\begin{tikzpicture}
  \node[inner sep=0pt] (sub_clusters) at (0,0) {\includegraphics[width=.22\textwidth]{sample_sub_cluster}};
  \node[inner sep=0pt, right = .1cm of sub_clusters ] (cooccur) {\includegraphics[width=.25\textwidth]{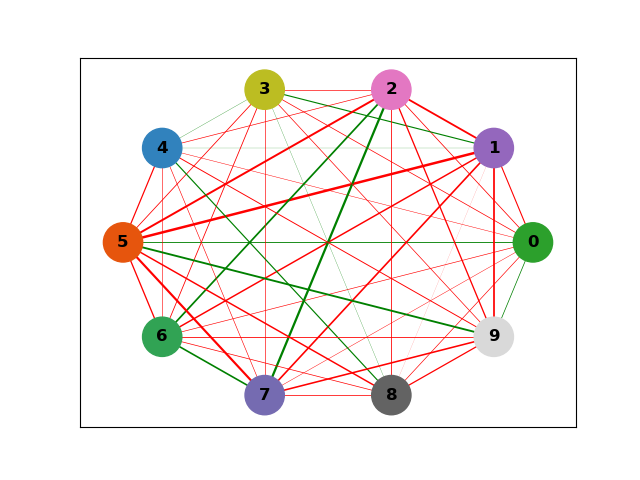}};
  \node[inner sep=0pt, below = .05cm of cooccur ] (seg_graph) {\includegraphics[width=.25\textwidth]{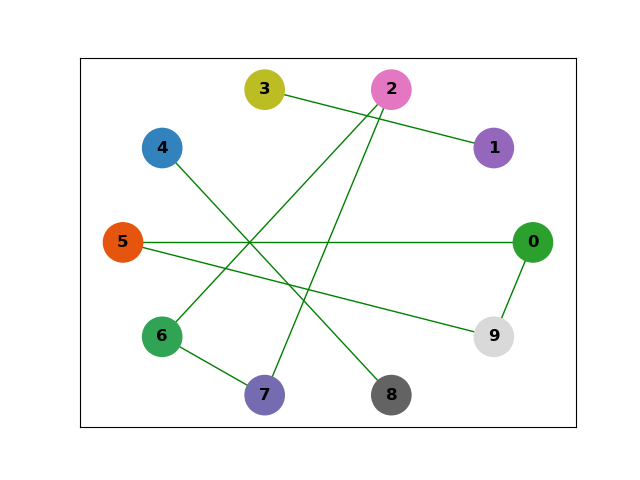}};
  \node[inner sep=0pt, left = .1cm of seg_graph ] (seg_inst) {\includegraphics[width=.22\textwidth]{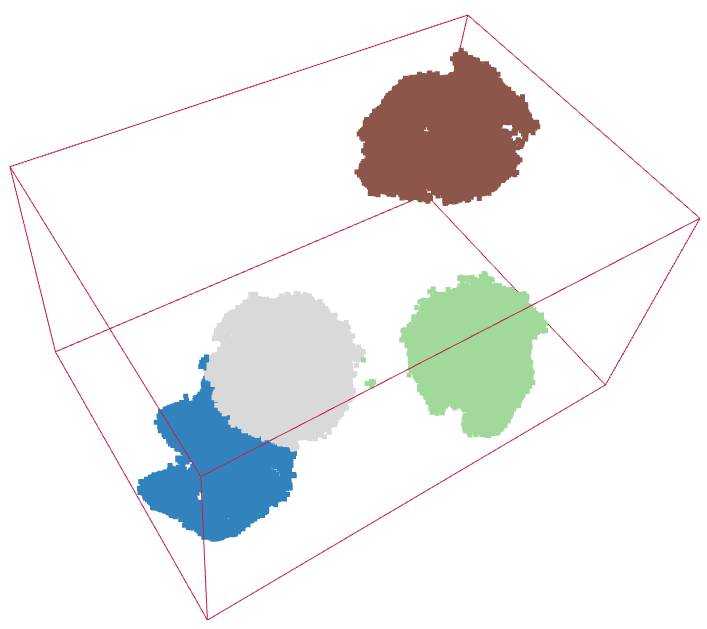}};
  \draw[->,thick] ($(sub_clusters.east)-(.2,0)$) -- ($(cooccur.west)+(.4,0)$);
  \draw[->,thick] ($(cooccur.south)+(0,.3)$) -- ($(seg_graph.north) - (0,.3)$);
  \draw[->,thick] ($(seg_graph.west)+(.45,0)$) -- ($(seg_inst.east)-(.2,0)$);
\end{tikzpicture}
\caption{The subcluster merging process. Top left: a single supercluster is
partitioned into $10$ subclusters, each represented by a unique color. Top
right: a weighted complete graph illustrates the affinities among subclusters,
with positive and negative affinities colored in green and red, respectively.
The width of each edge is proportional to its affinity value. Bottom right: the
graph is partitioned into smaller subgraphs based on affinity scores via label
propagation. Bottom left: the corresponding subclusters belonging to the same
subgraph are merged.}
\label{fig:merging_subclusters}
\end{figure}

The merging step aims to combine subclusters that constitute the same crop
instance. Ideally, for a given view $C_j$, we would merge a pair of subclusters
$S_i$ and $S_{i^{\prime}}$ if they share the same instance label in $M_j$.
However, we have $n$ such observations from $n$ corresponding views. Therefore,
to incorporate all the observations we introduce a subcluster affinity score
$\alpha_{ii^{\prime}}$ between $S_i$ and $S_{i^{\prime}}$. The score is
determined based on the mask reliability scores across $n$ views by
\begin{equation}
  \alpha_{ii^\prime} = \sum_{j=1}^{n} r_{ij}r_{i^{\prime}j} \cdot (-1)^{\mathds{1} \{\lambda_{ij}\ne \lambda_{i^{\prime}j}\}},
\end{equation}
where $\mathds{1}\{\cdot\}$ is an indicator function that evaluates to 1 when
the condition is satisfied and 0 otherwise. As shown in
Fig.~\ref{fig:merging_subclusters}, for each pair of subclusters in a
supercluster we compute the affinity score and build a weighted complete graph.
Each subcluster maps to a node in the graph and the affinity score between a
pair of subclusters is represented by the corresponding edge weights. Finally,
we apply a label propagation algorithm \cite{raghavan2007near} on the graph to
partition it into subgraphs that represent the individual crop instances.

\section{Evaluation}
\label{sec:evaluation}
\subsection{Cotton Plant Dataset}
\begin{figure}
\centering
\begin{minipage}{.47\linewidth}
  \centering
  \includegraphics[width=\textwidth, height=5.3cm,  trim={0cm 0cm 0cm 0cm},clip] {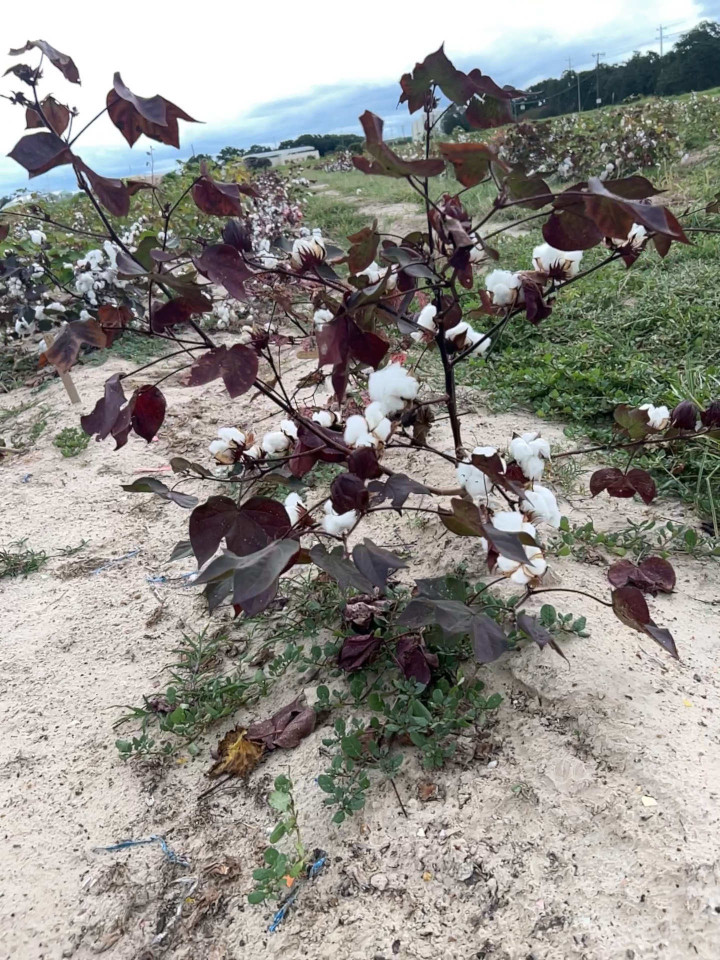}
  \subcaption{Cotton plant.}
\end{minipage}
\begin{minipage}{.47\linewidth}
  \centering
  \includegraphics[width=.48\textwidth,]{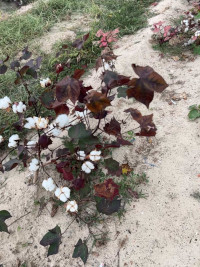}
  \includegraphics[width=.48\textwidth]{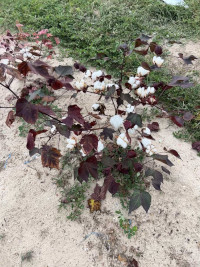}
  \vspace{-9pt}\\
  \includegraphics[width=.48\textwidth,]{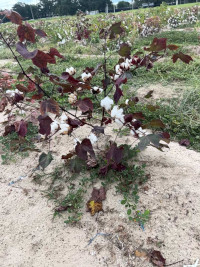} 
  \includegraphics[width=.48\textwidth]{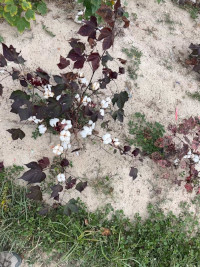} 
  \subcaption{Sample views.}
\end{minipage}
\caption{Example images from our cotton plant dataset. (a) shows a cotton plant
while (b) displays sample views captured from different viewpoints.}
\label{fig:dataset_samples}
\end{figure}

We collected a cotton plant dataset consisting of 8 plants recorded at the
Texas A\&M University Research Farm. The images were captured using an Apple
iPhone at a resolution of 1040$\times$1920 pixels. Approximately 150 images per
plant were taken from a distance of 1 m by recording multiple viewpoints.
Fig.~\ref{fig:dataset_samples} depicts representative samples of the camera
viewpoints during the data acquisition process. To estimate the image poses, we
utilized the Spectacular AI \cite{spectacular} mobile application. The
ground-truth counts for each plant were obtained by manually counting all
cotton bolls through direct observation. To generate instance masks we employed
the Segment Anything Model (SAM) \cite{kirillov2023segment}, a pretrained 2D
instance segmentation model, using manual bounding box supervision as input
prompts.

\subsection{Experimental Setup}
In addition to evaluating CropNeRF on our cotton plant dataset, we assessed its
performance on apple and pear tree data provided by FruitNeRF. The apple tree
dataset consists of 3 trees, each photographed using a DSLR camera with a 35 mm
lens, producing images at a resolution of 4000$\times$6000 pixels. Roughly 350
images per tree were captured from an approximate distance of 3 m, covering
various angles and heights. COLMAP \cite{schonberger2016structure} was used to
estimate the camera poses. The pear tree dataset was synthetically generated
using Blender \cite{blender}. The virtual camera was configured with a 35 mm
focal length and an image resolution of 1024$\times$1024 pixels. Among the
different types of crop data, cotton presents a greater challenge. This is due
to the considerable variation in the size and shape of cotton bolls in contrast
to the more uniform dimensions of apples and pears. Consequently, our
evaluation primarily focuses on cotton. 

\subsection{Implementation Details}
For the 3D reconstruction and point cloud extraction, we made use of existing
NeRF models. Specifically, CropNeRF is based on Nerfacto
\cite{tancik2023nerfstudio}, augmented with a semantic field, and follows the
architectural details of FruitNeRF. Given the differences in size among the
cotton plants and apple and pear trees, we employed the FruitNeRF network to
render the cotton plants while the FruitNeRF-Big network was used to render the
apple and pear trees. The models were trained for 30,000 and 100,000 epochs,
respectively. To train our model with semantic masks, we derived the masks from
the instance masks generated via SAM by assigning a uniform label to all
instances. CropNeRF was trained on an Ubuntu 22.04.5 LTS machine with an Intel
Xeon Gold 6330 2.00 GHz CPU with 64 GB of memory and an NVIDIA A100 GPU.
Training each cotton plant model required about 12 minutes.

To partition the exported point cloud into superclusters, we applied DBSCAN with
parameters $eps=0.02$ and $min\_points=30$. K-means clustering was performed in
the second phase to partition each supercluster into $K$ subclusters where
$K=10$. Notably, the same parameter settings were used for both datasets. In
Sec.~\ref{subsec:k_vs_count}, we demonstrate the robustness of CropNeRF by
analyzing the relationship between the number of subclusters and the associated
counts. To compute the occlusion-aware projection
$\mathcal{V}_j\left(S_i\right)$ in \eqref{eq:visibility_score} and
\eqref{eq:consistency_score} we employed a z-buffering technique, which is
commonly used in computer graphics. The tree-level count was obtained by summing
the individual instance counts across all superclusters.

\subsection{Counting Accuracy}
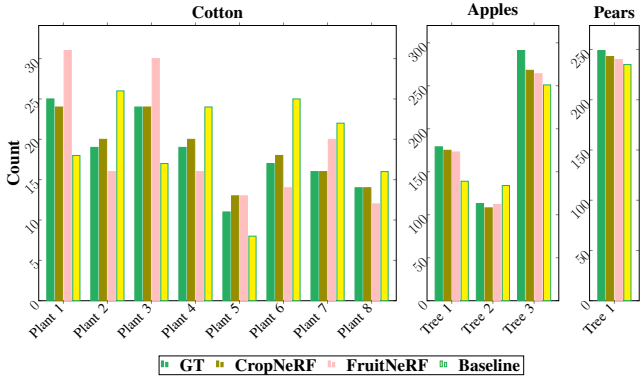
\begin{figure}
\centering
\resizebox{\linewidth}{!}{

\definecolor{bblue}{HTML}{4F81BD}
\definecolor{rred}{HTML}{85c1e9}
\definecolor{ggreen}{HTML}{27ae60}
\definecolor{ppurple}{HTML}{884ea0}

\pgfplotsset{every axis/.append style={
                    label style={font=\Huge},
                    tick label style={font=\Huge}
                    }}

\begin{tikzpicture}
\begin{groupplot}[
    group style={
        group size=3 by 1,
        horizontal sep=1.5cm,
        ylabels at=edge left,
    },
    ylabel={Count},
    label style={font=\bfseries\Huge},
    title style={font=\bfseries\Huge},       
    width=5cm,
    height=15cm,
    ybar,
    ymin=0,
    enlarge x limits=0.2,
    xtick=data,
    ticklabel style={font=\bfseries\Huge, rotate=45, anchor=east},
    legend cell align=left,
    legend style={
        at={(-5,-.2)},
        anchor=north,
        legend columns=-1,
        column sep=.5cm,
        font=\bfseries\Huge
    }
]

\nextgroupplot[title={Cotton},ytick distance=5,tick label style={font=\huge}, width=19cm, symbolic x coords={Plant 1,Plant 2,Plant 3,Plant 4,Plant 5,Plant 6,Plant 7,Plant 8}, enlarge x limits=0.08]
\addplot[style={ggreen,fill=ggreen,mark=none}] coordinates {(Plant 1, 25) (Plant 2, 19) (Plant 3, 24) (Plant 4, 19) (Plant 5,11) (Plant 6,17)
        (Plant 7, 16) (Plant 8, 14) };

        \addplot[style={olive,fill=olive,mark=none}]
             coordinates {(Plant 1,24) (Plant 2,20) (Plant 3,24) (Plant 4,20) (Plant 5,13) (Plant 6,18)
        (Plant 7,16) (Plant 8,14) };
        \addplot[style={pink,fill=pink,mark=none}]
             coordinates {(Plant 1,31) (Plant 2,16) (Plant 3,30)(Plant 4,16) (Plant 5,13) (Plant 6,14)
        (Plant 7,20) (Plant 8,12) };

         \addplot[style={ggreen,fill=yellow,mark=none}]
            coordinates {(Plant 1, 18) (Plant 2, 26) (Plant 3, 17) (Plant 4, 24) (Plant 5,8) (Plant 6,25)
        (Plant 7, 22) (Plant 8, 16) };
\nextgroupplot[title={Apples}, ytick distance=50, tick label style={font=\huge}, width=8cm, symbolic x coords={Tree 1,Tree 2,Tree 3}, enlarge x limits=0.3]

 \addplot[style={ggreen,fill=ggreen,mark=none}]coordinates {(Tree 1, 179) (Tree 2, 113) (Tree 3, 291)};
 \addplot[style={olive,fill=olive,mark=none}]coordinates {(Tree 1,175) (Tree 2,108) (Tree 3,268) };
 \addplot[style={pink,fill=pink,mark=none}]coordinates {(Tree 1,173) (Tree 2,112) (Tree 3,264)};
 \addplot[style={ggreen,fill=yellow,mark=none}]coordinates {(Tree 1, 139) (Tree 2, 134) (Tree 3, 251)};

\nextgroupplot[title={Pears}, ytick distance=50, tick label style={font=\huge}, width=4cm, symbolic x coords={Tree 1}, enlarge x limits=0.1]
\addplot[style={ggreen,fill=ggreen,mark=none}]
            coordinates {(Tree 1, 249) };

        \addplot[style={olive,fill=olive,mark=none}]
             coordinates {(Tree 1,243) };

        \addplot[style={pink,fill=pink,mark=none}]
             coordinates {(Tree 1,240) };
         \addplot[style={ggreen,fill=yellow,mark=none}]
            coordinates {(Tree 1, 235)};
\legend{GT, CropNeRF, FruitNeRF, Baseline}
\end{groupplot}
\end{tikzpicture}
}
\caption{A cotton boll, apple, and pear counting comparison involving the
ground truth (GT), CropNeRF, FruitNeRF, and the baseline.}
\label{fig:counting_comparison}
\vspace{-2mm}
\end{figure}

CropNeRF was tested against a baseline method and recent 3D rendering crop
counting approaches using the same input segmentation masks. The baseline
technique generates point clouds using COLMAP through an SfM algorithm, followed
by segmentation using the clustering method employed in FruitNeRF.
Fig.~\ref{fig:counting_comparison} illustrates the per-plant/tree crop counts.
For the FruitNeRF results on the apple and pear datasets, we reported the values
published in the original study, which used individually tuned per-tree
parameter settings. To evaluate FruitNeRF on the cotton dataset, the
$template\_size$ parameter was set to 30 mm to be consistent with the average
cotton boll diameter reported by Wallace and Fields \cite{wallace1997boll}. We
further evaluated CropNeRF using multiple counting error metrics, which are
tabulated in Table~\ref{tab:count_error_comparison}. 

In comparison to FruitNeRF, CropNeRF consistently demonstrated superior
performance. Specifically, CropNeRF achieved a mean absolute percentage error
(MAPE) of 4.9\% for cotton bolls and 4.7\% for apples, indicating a consistent
level of accuracy across diverse crop types. On the other hand, FruitNeRF
yielded a MAPE of 18.1\% for cotton bolls and 4.5\% for apples, revealing a
significant variation in performance between the two crops. This deviation
stems from the dependency of clustering-based point cloud segmentation methods
on predefined template crop sizes and shapes. For pears, both frameworks
achieved relatively low MAPE values (2.4\% for CropNeRF versus 3.6\% for
FruitNeRF) compared to apples and cotton bolls, which can be attributed to the
fact that pears typically grow individually rather than in dense clusters.
Compared to Cotton3DGaussians \cite{jiang2025cotton3dgaussians}, CropNeRF
produced lower counting errors despite operating in a real-world outdoor
environment. These results highlight the robustness and versatility of CropNeRF
for counting tasks across different agricultural commodities.

\begin{table}
\centering
\begin{tabular}{c|c|c|c|c|c|c}
\hline
\textbf{Method} & \multicolumn{2}{c}{Cotton Bolls} & \multicolumn{2}{c}{Apples} & \multicolumn{2}{c}{Pears}\\
\hline
& \textbf{R} & \textbf{M}
& \textbf{R} & \textbf{M}
& \textbf{R} & \textbf{M}\\
\hline
Baseline                                            & 5.9          & 30.8         & 34.8          & 18.2         & 14.0         & 5.6 \\
FruitNeRF \cite{meyer2024fruitnerf}                 & 3.9          & 18.1         & 15.9          & \textbf{4.5} & 9            & 3.6 \\
Cotton3DGaussians \cite{jiang2025cotton3dgaussians} & 1.7          & 9.2          & -             & -            & -            & - \\
CropNeRF (ours)                                     & \textbf{1.0} & \textbf{4.9} & \textbf{13.7} & 4.7          & \textbf{6.0} & \textbf{2.4} \\
\hline
\end{tabular}
\caption{A comparison of counting errors across different methods. \textbf{R}
stands for root mean squared error and \textbf{M} signifies mean absolute
percentage error.}
\label{tab:count_error_comparison}
\end{table}


\subsection{Mask Discrepancy Handling}
\begin{figure}
\centering
\begin{subfigure}[]{0.24\linewidth}
  \includegraphics[width=\textwidth, height=1.8cm, trim={0cm 0cm 0cm 0cm},clip]{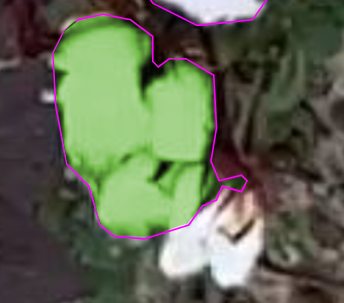}
  \caption{}
  \label{fig:mask_mismatch}
\end{subfigure}
\begin{subfigure}[]{0.24\linewidth}
  \includegraphics[width=\textwidth, height=1.8cm, trim={0cm 0cm 0cm 0cm},clip]{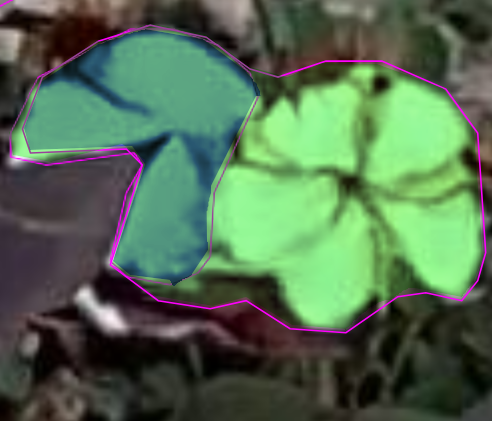}
  \caption{}
  \label{fig:instance_merged1}
\end{subfigure}
\begin{subfigure}[]{0.24\linewidth}
  \includegraphics[width=\textwidth,height=1.8cm, trim={0cm 0cm 0cm 0cm},clip]{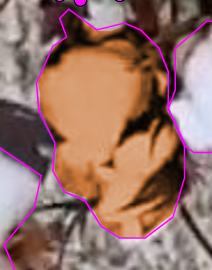}
  \caption{}
  \label{fig:instance_merged2}
\end{subfigure}
\begin{subfigure}[]{0.24\linewidth}
  \includegraphics[width=\textwidth, height=1.8cm, trim={0cm 0cm 0cm 0cm},clip]{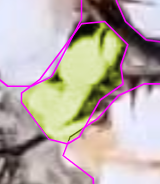}
  \caption{}
  \label{fig:instance_merged3}
\end{subfigure}
\begin{subfigure}[]{0.24\linewidth}
  \includegraphics[width=\textwidth, height=1.8cm, trim={0cm 0cm 0cm 0cm},clip]{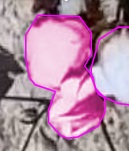}
  \caption{}
  \label{fig:instance_merged4}
\end{subfigure}
\begin{subfigure}[]{0.24\linewidth}
  \includegraphics[width=\textwidth, height=1.8cm, trim={0cm 0cm 0cm 0cm},clip]{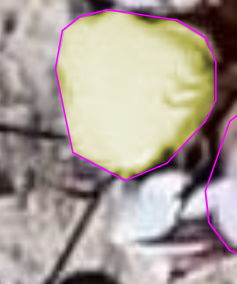}
  \caption{}
  \label{fig:misdetected}
\end{subfigure}
\begin{subfigure}[]{0.24\linewidth}
  \includegraphics[width=\textwidth, height=1.8cm, trim={0cm 0cm 0cm 0cm},clip]{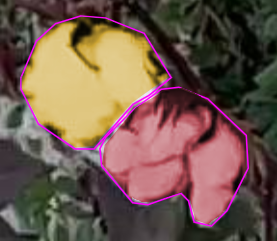}
  \caption{}
  \label{fig:correct_mask}
\end{subfigure}
\begin{subfigure}[]{0.24\linewidth}
  \includegraphics[width=\textwidth, height=1.8cm, trim={0cm 0cm 0cm 0cm},clip]{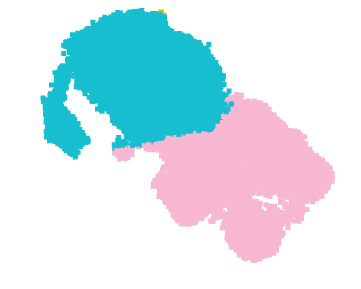}
  \caption{}
  \label{fig:inconsistency_outcome}
\end{subfigure}
\caption{Examples of mask inconsistencies where two cotton boll instances are
inconsistently labeled across multiple views: (a) the mask does not align with
the corresponding instance boundary; (b-e) two bolls are incorrectly labeled as
a single instance; (f) one of the bolls is missed by the detector; (g) bolls are
correctly labeled; (h) CropNeRF's reliability score attenuates inconsistencies
and allows for successful identification of the bolls.}
\label{fig:mask_inconsistencies}
\vspace{-2mm}
\end{figure}

As displayed in Fig.~\ref{fig:mask_mismatch}, instance masks generated by SAM
can exhibit inconsistencies, i.e., they do not fully align with the
corresponding instances. Additionally, multiple distinct instances may be
erroneously labeled as a single instance (e.g.,
Fig.~\ref{fig:instance_merged1}-\ref{fig:instance_merged4}), while in other
cases certain instances may remain unlabeled due to misdetection (e.g.,
Fig.~\ref{fig:misdetected}). These discrepancies are especially evident in the
cotton plant dataset, where the irregular shape of the bolls makes it
challenging to differentiate individual instances precisely.

CropNeRF is able to mitigate many of these errors and successfully identify crop
instances, Fig.~\ref{fig:inconsistency_outcome}. For instance, the reliability
score is low in scenarios such as Fig.~\ref{fig:mask_mismatch} and
Fig.~\ref{fig:misdetected} due to low mask consistency. Similarly, scenarios
such as Fig.~\ref{fig:instance_merged2}, Fig.~\ref{fig:instance_merged3}, and
Fig.~\ref{fig:instance_merged4} yield lower reliability scores due to reduced
visibility scores. With lower reliability scores these scenarios have a
diminishing effect on subsequent affinity calculations, resulting in
segmentations that are robust to mask discrepancies.

\subsection{Instance Mask Effectiveness}
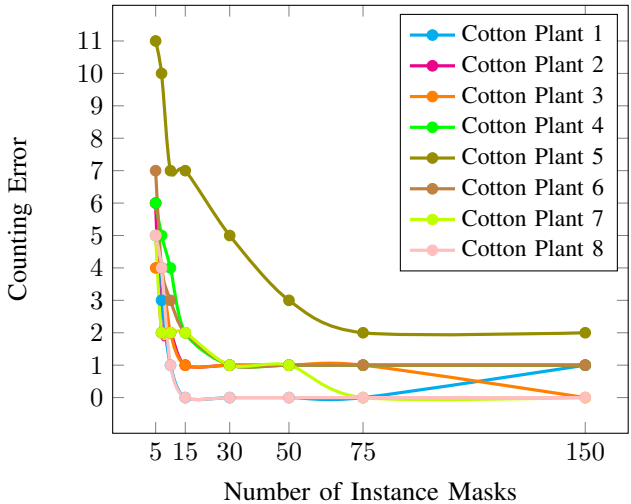
\begin{figure}
\centering
\pgfplotsset{
every axis plot/.append style={line width=1.2pt},
every axis plot post/.append style={
every mark/.append style={line width=.4pt,mark=*}
}
}
\begin{tikzpicture}
  \begin{axis}[
    x tick label style={rotate=0},
    legend style={font=\small,legend cell align=left},
    xtick={5,15,30,50,75,150}, 
    ytick={0,1,2,3,4,5,6,7,8,9,10,11},
    xlabel= Number of Instance Masks,
    ylabel = Counting Error,]
    \addlegendentry{Cotton Plant 1};
    \addplot[color=cyan,smooth,mark=.] table [x=n_view, y=t1_error, col sep=comma] {data/view_vs_error.csv};
    \addlegendentry{Cotton Plant 2};
    \addplot[color=magenta,smooth,mark=.] table [x=n_view, y=t2_error, col sep=comma] {data/view_vs_error.csv};
    \addlegendentry{Cotton Plant 3};
    \addplot[color=orange,smooth,mark=*] table [x=n_view, y=t3_error, col sep=comma] {data/view_vs_error.csv};
    \addlegendentry{Cotton Plant 4};
    \addplot[color=green,smooth,mark=.] table [x=n_view, y=t4_error, col sep=comma] {data/view_vs_error.csv};
    \addlegendentry{Cotton Plant 5};
    \addplot[color=olive,smooth,mark=.] table [x=n_view, y=t5_error, col sep=comma] {data/view_vs_error.csv};
    \addlegendentry{Cotton Plant 6};
    \addplot[color=brown,smooth,mark=.] table [x=n_view, y=t6_error, col sep=comma] {data/view_vs_error.csv};
    \addlegendentry{Cotton Plant 7};
    \addplot[color=lime,smooth,mark=.] table [x=n_view, y=t7_error, col sep=comma] {data/view_vs_error.csv};
    \addlegendentry{Cotton Plant 8};
    \addplot[color=pink,smooth,mark=.] table [x=n_view, y=t8_error, col sep=comma] {data/view_vs_error.csv};
    \end{axis}
\end{tikzpicture}
\caption{The number of instance masks versus the counting error.}
\label{fig:number_of_masks_vs_counting_error}
\end{figure}

We evaluated the effectiveness of CropNeRF on counting tasks using a sparse set
of instance masks. In this experiment, we varied the number of available masks
$m$ from 5 to 150 and analyzed the resulting boll counting error across
different cotton plants. For each value of $m$, a set of $m$ masks is uniformly
sampled from the complete set of 150 masks. The results presented in
Fig.~\ref{fig:number_of_masks_vs_counting_error} indicate that in most cases
only 15 to 30 masks are sufficient to achieve an accurate count.

\subsection{Number of Subclusters Impact}
\label{subsec:k_vs_count}
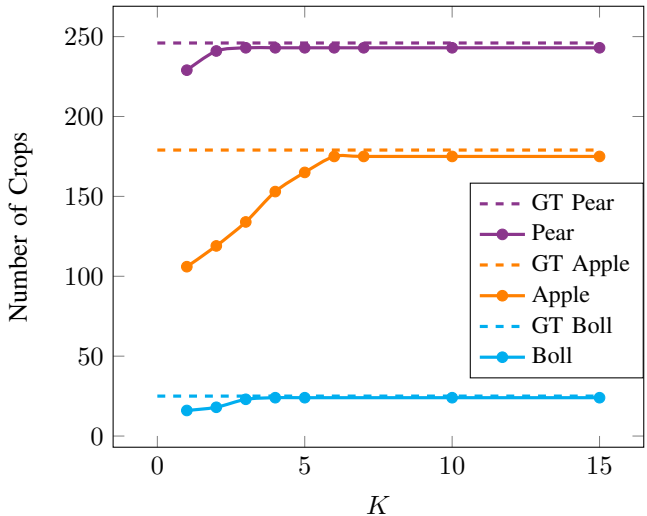
\begin{figure}
\centering
\pgfplotsset{
  every axis plot/.append style={line width=1.2pt},
  every axis plot post/.append style={
    every mark/.append style={line width=.4pt,mark=*}
  }
}
\begin{tikzpicture}
  \begin{axis}[
    width=\linewidth,
    legend style={at={(1,.6)},anchor=north east},
    x tick label style={rotate=0},
    legend style={font=\small,legend cell align=left},
    restrict x to domain=0:150,
    xlabel= $K$,
    ylabel = Number of Crops,
  ]
  \addlegendentry{GT Pear};
  \addplot[mark=none, Fuchsia, dashed]coordinates {(0,246)(15,246)};
  \addlegendentry{Pear};
  \addplot[color=Fuchsia,smooth,mark=.]coordinates {(1,229) (2,241) (3,243) (4,243) (5,243) (6,243) (7,243) (10,243) (15,243)};

  \addlegendentry{GT Apple};
  \addplot[mark=none, orange, dashed]coordinates {(0,179)(15,179)};
  \addlegendentry{Apple};
  \addplot[color=orange,smooth,mark=.]coordinates {(1,106) (2,119) (3,134) (4,153) (5,165) (6,175) (7,175) (10,175) (15,175)};
  \addlegendentry{GT Boll};
  \addplot[mark=none, cyan, dashed] coordinates {(0,25) (15,25)};
  \addlegendentry{Boll};
  \addplot[color=cyan,smooth,mark=.]coordinates {(1,16) (2,18) (3,23) (4,24) (5,24) (10,24) (15,24)};
  \end{axis}
\end{tikzpicture}
\caption{The number of subclusters ($K$) versus the crop count. The dashed line
represents the ground-truth (GT) count while the solid line indicates the number
of crops identified by CropNeRF.}
\label{fig:k_vs_count}
\end{figure}

A primary goal in designing CropNeRF was to minimize the need for crop-specific
parameter optimization. Accordingly, we analyzed the relationship between the
parameter $K$, a key input to the k-means clustering algorithm, and the counting
accuracy. Samples from the apple tree, pear tree, and cotton plant datasets were
selected to evaluate the influence of $K$. The results are plotted in
Fig.~\ref{fig:k_vs_count}.

For cotton bolls, the results show that the count remained consistent for $K
\ge 4$. This stability arises from the fact that in most cases the number of
instances within a supercluster does not exceed five. The maximum number of
bolls observed within a single supercluster was four for the cotton plant
dataset. Similarly, for apples the count stayed approximately constant with $K
\ge 6$ as the apple clusters typically contained a maximum of six to seven
instances. In the case of pears, clusters generally comprised only one or two
pears, leading to stable counts for $K \ge 2$. 

Based on these empirical findings, we set $K = 10$ for all the experiments. In
general, we recommend selecting $K$ to be greater than the expected number of
instances within a cluster. Nevertheless, $K$ should not be excessively large
since this will increase computational complexity due to the higher number of
subclusters.

\subsection{Ablation Study}
\begin{table}
\centering
\begin{tabular}{p{2cm}|p{1cm}p{1cm}p{1cm}|c}
  \hline
  \textbf{Method}       & \textbf{Visibility} & \textbf{Mask}       & \textbf{LPA} & \textbf{MAPE\%\textcolor{green}{$\downarrow $}} \\\hline
  Baseline              &                     &                     &              & 7.1 \\ \hline
  $+$visibility         & \checkmark          &                     &              & 6.3 \\ \hline
  $+$mask               &                     & \checkmark          &              & 6.5 \\ \hline
  $+$mask,$+$visibility & \checkmark          & \checkmark          &              & 5.4 \\ \hline
  CropNeRF              & \checkmark          & \checkmark          & \checkmark   & \textbf{4.9} \\ \hline
\end{tabular}%
\caption{An ablation study on the cotton plant dataset. The first row is the
baseline method, which merges subclusters using only the instance mask. In
subsequent rows, the integration of visibility and mask consistency scores
progressively enhances the counting accuracy. The last row, representing
CropNeRF, applies a label propagation algorithm (LPA) in addition to visibility
and mask consistency scores resulting in the lowest error.}
\label{tab:ablation_study}
\end{table}

To understand the impact of key design choices in CropNeRF, we examined the
following components: visibility score, mask consistency score, and label
propagation algorithm. The ablation study results are presented in
Table~\ref{tab:ablation_study} and described as follows. The baseline method
omits the visibility and mask consistency scores. Instead, it merges subcluster
pairs if and only if $\alpha_{ii^\prime} > 0$, where $\alpha_{ii^\prime} =
\sum_{j=1}^{n} (-1)^{\mathds{1} \{\lambda_{ij}\ne \lambda_{i^{\prime}j}\}}$.
The $+$visible and $+$mask methods, shown in the second and third rows
respectively, compute an affinity score based on either the visibility score
$\alpha_{ii^\prime} = \sum_{j=1}^{n} v_{ij} \cdot (-1)^{\mathds{1}
\{\lambda_{ij}\ne \lambda_{i^{\prime}j}\}}$ or the mask consistency score
$\alpha_{ii^\prime} = \sum_{j=1}^{n} c_{ij} \cdot (-1)^{\mathds{1}
\{\lambda_{ij}\ne \lambda_{i^{\prime}j}\}}$. Subcluster pairs are merged if and
only if $\alpha_{ii^\prime} > 0$.

CropNeRF incorporates both visibility and mask consistency scores to compute a
reliability score along with a label propagation strategy for subcluster
merging. The ablation study results highlight that the visibility score is the
dominant factor in determining segmentation performance followed by the mask
consistency score. Consequently, when affinity scores are derived from these two
factors, merging if and only if $\alpha_{ii^\prime}>0$ is sufficient for
achieving a precise segmentation.

%
\section{Limitations and Future Work}
\label{sec:conclusion_and_future_work}
Although NeRFs and their variants can produce high-fidelity 3D representations,
they require substantial computing resources. In the CropNeRF pipeline, NeRF
model training is the most computationally intensive and time-consuming stage.
Furthermore, accurately detecting certain crops within dense canopy structures
remains difficult due to the spectral limitations of RGB cameras. Future work
will not only focus on substitutes to NeRFs, but also on integrating additional
sensing modalities such as normalized difference vegetation index and thermal
imaging to enhance crop detection and improve the overall robustness of the
framework.

\section{Conclusion}
This paper introduced CropNeRF, an occlusion-aware counting framework that
accurately enumerates individual crop instances via 3D instance segmentation.
Furthermore, we created a multi-view cotton plant image dataset to promote
future research in this area. CropNeRF is highly robust and does not require
sensitive parameter tuning for different crop types. We evaluated the
effectiveness of CropNeRF on three distinct crop datasets (cotton bolls,
apples, and pears), which drastically vary in color, shape, and size.
Experimental results demonstrate accurate counts that exceed the state of the
art across all three datasets, highlighting framework's adaptability and
reliability for agricultural applications.

\section*{Acknowledgments}
Md Ahmed Al Muzaddid was supported by a University of Texas at Arlington
Dissertation Fellowship. We thank Aaron J. DeSalvio for assistance with the
data collection at Texas A\&M University. We acknowledge the Texas Advanced
Computing Center (TACC) at The University of Texas at Austin for providing
software, computational, and storage resources that have contributed to these
results.

\bibliographystyle{IEEEtran}
\bibliography{IEEEabrv,cropnerf}

\end{document}